          [2001/04/25 1.1 (PWD)]
\documentclass[a4paper,twocolumn]{esapub2005} 
\usepackage{times}
\usepackage[numbers]{natbib}
\usepackage{url}
\usepackage{siunitx}
\usepackage{svg}
\usepackage{bbm}
\usepackage{subcaption}
\usepackage{amssymb}
\usepackage{amsmath}
\usepackage{hyperref}

\title{AUV trajectory optimization with hydrodynamic forces for
Icy Moon Exploration}
\author[1,2]{Lukas Rust}
\author[2,3]{Shubham Vyas}
\author[2]{Bilal Wehbe}
\author[1,2,3]{Frank Kirchner}
\affil[1]{Fachbereich 03, Universität Bremen, Germany, lrust@uni-bremen.de}
\affil[2]{Robotics Innovation Centre (RIC), DFKI GmbH, Germany, \{lukas.rust, bila.wehbe, shubham.vyas, frank.kirchner\}@dfki.de}
\affil[3]{AG Robotik, University of Bremen, Germany}

\newcommand{\btx}{\textsc{Bib}\TeX}
\newcommand{\filename}{esapub}

\usepackage{graphicx}
\usepackage{amsmath}
\usepackage{todonotes}

\begin{document}

\maketitle

\begin{abstract}
To explore oceans on ice-covered moons in the solar system, energy-efficient Autonomous Underwater Vehicles (\textit{AUVs}) with long ranges must cover enough distance to record and collect enough data. These usually underactuated vehicles are hard to control when performing tasks such as vertical docking or the inspection of vertical walls. This paper introduces a control strategy for \textit{DeepLeng} to navigate in the ice-covered ocean of \textit{Jupiter's} moon \textit{Europa} and presents simulation results preceding a discussion on what is further needed for robust control during the mission.
\end{abstract}

\section{Introduction}
In recent decades, many missions \cite{clark2009return,blanc2009laplace} have been undertaken to investigate Europa, one of Jupiter's moons. Its surface consists of a thick ice crust which covers a deep ocean underneath it. Even today multiple missions like \textit{ESA's} \textit{JUICE} \cite{grasset2013jupiter} mission or \textit{Clipper} by \textit{NASA} \cite{howell2020nasa} are prepared to launch to specifically orbit around Europa.
It attracts research because it is one of the most promising places in our solar system that could contain life\cite{chyba2000energy}.\\
To inspect the ocean a system was conceptualized at \textit{DFKI Bremen} which could drill through the ice and then deploy an \textit{AUV} into the ocean for inspection. The so-called \textit{DeepLeng} was built with design restrictions to minimize energy consumption and also spatial requirements to fit into the cargo space\cite{hildebrandt_design_2013}. The slender torpedo-like shape minimizes drag during forward locomotion, increasing the range and speed of the \textit{AUV}, which is necessary to realize long-term navigation missions to explore large oceans like \textit{Europa's}. The manifold advantages of such a design come at the cost of underactuation, thus the incapability to generate any wrench.  
The recently emerging field for control of underactuated systems offers many techniques, but applying them to the underwater domain demands the extension of the system model by hydrodynamic effects and deals with free-floating platforms and a singularity-free representation of the orientation.
The interest to develop and control such \textit{AUVs} exceeds the exploration of oceans of icy moons and allows the deployment of a new class of underwater robots.
This paper contributes a Trajectory Optimization formulation to generate trajectories for chosen agile maneuvers, which explicitly respect the hydrostatic, and hydrodynamic effects and the mechanic constraints. To account for various sources of error, a time-varying Linear Quadratic Regulator is proposed to stabilize the trajectory during execution.
The control architecture will show that despite the constraints, maneuvers far outside of the scope of movements, for which the system was originally designed, are possible.
\begin{figure}[t!]
\centering
\includegraphics[width=\linewidth]{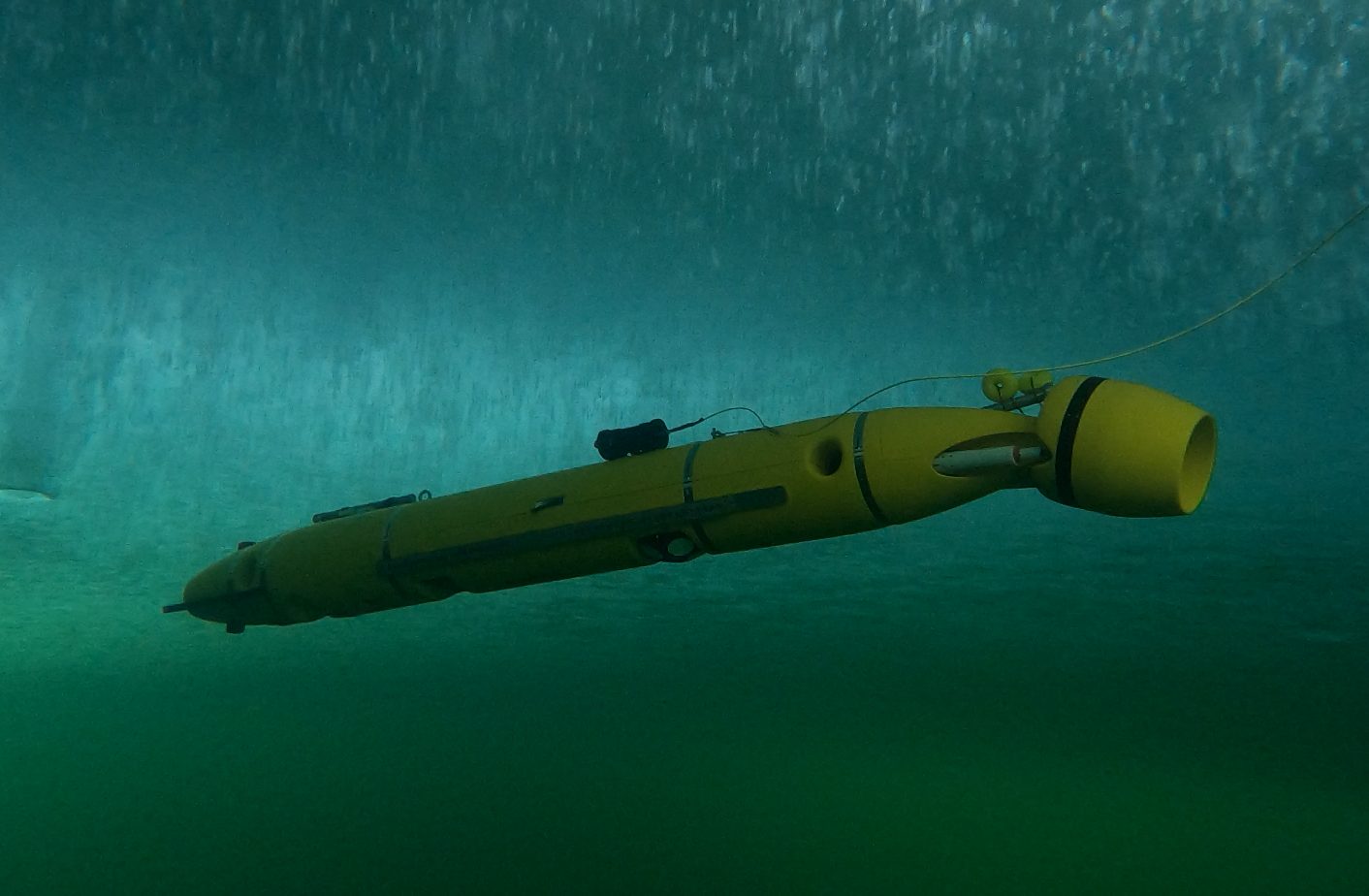}
\caption{\textit{DeepLeng AUV} demonstrating the vectored thrust during under-ice deployment in Abisko, Sweden}
\end{figure}
\subsection{Related Work} 
There is extensive research in the field of underactuated systems. Especially for canonical systems, like the cart-pole or the double pendulum, many approaches have been extensively studied. Besides Trajectory Optimization and Linear Quadratic Regulators (LQRs), Model Predictive Control (MPC) \cite{wagener2019online}, Reinforcement Learning Agents(RL) \cite{nagendra2017comparison} and Proportional Integrative Derivative (PID) \cite{yadav2011comparative} controllers have been used.\\
Usually, those canonical systems are fixed to the world. Underactuated free-floating platforms and their control were investigated to a lesser extent\cite {pei1998control, egeland2005free}. 
Trajectory Optimization in the underwater domain has been studied in \cite{aguiar2018trajectory} with a focus on long-distance navigation and not on agile maneuvering.
A comparable approach tested on a similar vehicle is presented in \cite{bhat2023controlling}. The featured \textit{AUV} is very similar to \textit{DeepLeng}. Both are only actuated by one large thruster at the rear of the vehicle. Nonlinear and linearized MPC controllers are used to perform hydrobatics such as the control of an AUV as an inverted pendulum. The controller can explicitly respect the system constraints posed by the maximum angle of the thruster while also respecting the hydrodynamic effects. Examples as sideward movements tested in the paper prove that the controller is capable of performing movements where classical techniques would generate actuation inputs that exceed the constraints of the system.\\
Due to the high computational cost, MPC can only optimize the control inputs for a restricted horizon and a reduced model. When generating reference trajectories offline or before the execution of the trajectory, one can use a nonlinear model over a longer horizon and then stabilize around the trajectory with the MPC from \cite{bhat2023controlling} or other approaches.

\section{System Dynamics} 
\subsection{DeepLeng} 
\textit{DeepLeng} has a single thruster at the rear of its body. It is mounted with a pan-tilt unit consisting of a rod and two linear actuators which fix the thruster to the main body and allow to alter the yaw and pitch angle of the thruster to the body.\begin{small}
\begin{figure}[htpb]
\centering
\includegraphics[width=\linewidth]{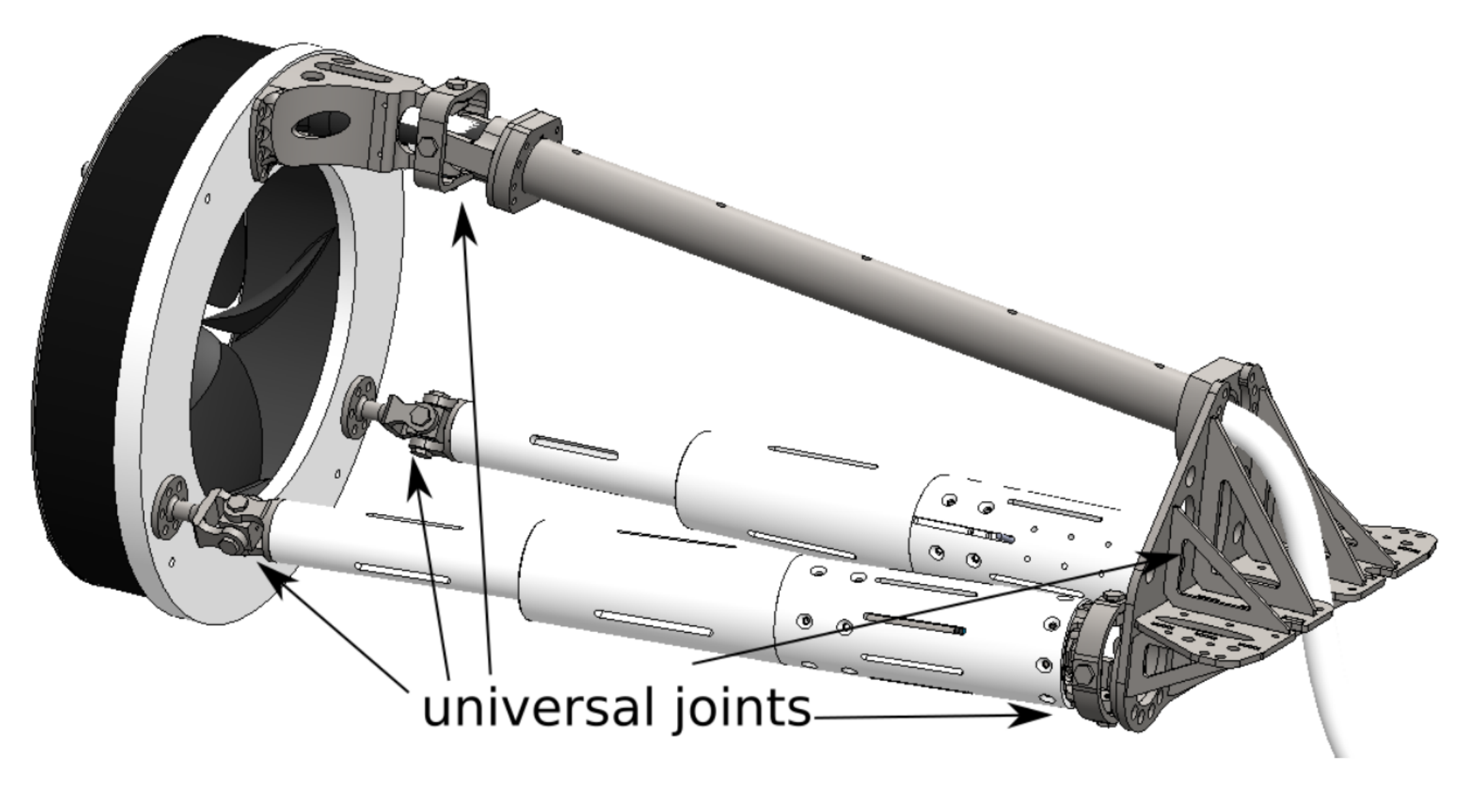}
\caption{Pan-tilt unit of \textit{DeepLeng}\label{fig:vector_thruster}}
\end{figure}
\end{small}The mechanism offers high stiffness and robustness, but the maximum angular speed of the thruster is relatively slow.\\
Following the parallel kinematics of this mechanism are not considered. Instead, the yaw and pitch angle of the thruster relative to the main body are used. Solutions to the inverse kinematics $(l_1, l_2) = k(\psi,\phi)$ of this mechanism which maps the thruster angles to the length of the linear actuators are well known and can be computed online.
The relation between a wrench $\omega$ acting on the thruster and the generalized forces $f$ acting on the body is given by:
\begin{equation}
    f = J^T(s)\omega 
\end{equation}
Where $J^{8\times6}$ is the Jacobian of the rigid multibody kinematics and $s$ the generalized positions of the system. 

\subsection{Dynamics}  
Computing the full hydrodynamic effects is computationally expensive and not solvable online. To still estimate the force affecting the \textit{AUV} one has to use simplified models to compute the drag forces.\\
Fossen\cite{fossen2011handbook} models the hydrodynamics as:
\begin{equation}
    M\dot{v} C(v)v + D(v)v + r(s) = \tau(s,u)
\end{equation}
$M$ is the mass matrix consisting of the mass matrix of the vehicle and $M_{AM}$ of the added mass due to the water which must be accelerated with the body. $C(v) = C(v)_M + C_{AM}$ are the Coriolis and centripetal forces of the mass matrix and the added mass.\\
$D(v)v = D_lv+D_q|v|v$ is the damping term consisting of two coefficients representing linear and quadratic drag terms.\\
$r(s)$ are the restoring forces. Since we assume a neutrally buoyant vehicle, for the gravitational force and the buoyant force $|f_g| = |f_b|$ holds, it only has a torque component:
\begin{equation}
    c_g \times f_g - c_b \times f_b
\end{equation}
$c_g$ being the center of gravity and $c_b$ the center of bouyancy.

\section{Control}
\subsection{Trajectory Optimization}
The decision variables of the Trajectory Optimization are generated by using Direct Transcription. Defining a number of time steps $T \in \mathbb{N}$,  we define a set of states $X = \{x_0, ..., x_{T+1}\}$ and control inputs $U = \{u_0,..., u_T\}$. To allow the solver to find a time optimal solution, we define a decision variable $h_{\Delta}$ representing the time step, and the total length of the trajectory as $D = Th_{\Delta}$. The decision variables are $Z = \{X,U,d_{\Delta} \}$
The state derivative is defined as $\dot{s_{t}} = f(s_t,u_t)$, where $f(s_t,u_t)$ is the nonlinear state derivative at a state $s_t$ and an input $u_t$.\\
States and the input are defined as:
\begin{eqnarray}
s = \left(\begin{array}{c}q_w,q_x,q_y,q_z,x,y,z,\psi,\phi\end{array}\right)^T\\
v = \left(\begin{array}{c}w_x,w_y,w_z,\dot{x},\dot{y},\dot{z},\dot{\psi},\dot{\phi}\end{array}\right)^T\\
s_q = \left(\begin{array}{c}q_w,q_x,q_y,q_z\end{array}\right)^T\\
s_p = \left(\begin{array}{c}x,y,z\end{array}\right)^T\\
x = \left(\begin{array}{c}s\\v\end{array}\right)\\
u = \left(\begin{array}{c}f, \tau_\psi, \tau_\phi\end{array}\right)^T
\end{eqnarray}
The generalized positions $s$ consist of a quaternion and the three-dimensional position, the yaw angle $\psi$ and the pitch angle $\phi$ of the thruster. The corresponding velocity $v$ consists of three angular velocities around the three principal axes and the derivatives of the position and the thruster angles. $s_q$ is the quaternion part of the state vector, and $s_p$ is the position part.
$v$ is the derivative of $s$, except that the derivative of the orientation is given as angular velocities instead of derivatives of the quaternion.
$x$ is the state vector of the system.
The input vector contains the force applied by the thruster, whose direction and thus its resulting wrench on the main body depends on $\psi$ and $\phi$. $\tau_\psi$ and $\tau_\phi$
are the torques applied on the virtual joints through which the thruster is mounted to the main body.
The optimization problem:\\
\begin{equation}
\underset{S,V,u}{\text{min}} 
\sum_{t=0}^{T+1} \left(s_p^t - s_p^f \right)^2
+ 
\sum_{t=0}^{T} u_t^2
+ 
(Td_{\Delta})^2 \label{eq:costs}\\[15pt]
\end{equation}
Subject to:\\
\begin{eqnarray}
F_{min} \leq f \leq F_{max} \label{eq:ulimits}\\[5pt]
h_{\Delta_{min}} \leq h_\Delta \leq h_{\Delta_{max}} \label{eq:hlimit}\\[5pt]
0.999 \leq \lVert s_q \lVert \leq 1.001 \label{eq:quatnorm}\\[5pt]
\dot{s_q} = \left(\begin{array}{c}
0 \\
w_x \\
w_y \\
w_z
\end{array}\right) * s_q \label{eq:quatmult}\\[5pt]  
s_{t+1} = s_t + h_\Delta \dot{s}_{t+1} \label{eq:posint}\\[5pt]
v_{t+1} = v_t + h_\Delta \dot{v}_{t+1}\label{eq:velint}\\[5pt]
M \dot{v}_t + C v + D(v) v  = J^T \left(\begin{array}{c}
0 \\
\tau_\phi\\
\tau_\psi\\
f \\
0\\
0
\end{array}\right)\label{eq:maneqn}\\[5pt]
h_{\Delta_{t+1}} = h_{\Delta_{t}} \label{eq:hconst}\\[5pt]
\psi_{min} \leq \psi \leq \psi_{max}\label{eq:psi}\\[5pt]
\dot{\psi}_{min} \leq \dot{\psi} \leq \dot{\psi}_{max}\label{eq:psid}\\[5pt]
\phi_{min} \leq \phi \leq \phi_{max}\label{eq:phi}\\[5pt]
\dot{\phi}_{min} \leq \dot{\phi} \leq \dot{\phi}_{max}\label{eq:phid}\\[5pt]
s_{t=0} = s_{init}\\[5pt]
v_{t=0} = v_{init}\\[5pt]
\dot{v}_{t=0} = \dot{v}_{init}\\[5pt]
s_{t=T} = s_f\\[5pt]
v_{t=T} = v_{f}\\[5pt]
\dot{v}_{t=T} = \dot{v}_{f}
\end{eqnarray}
The costs are defined in \autoref{eq:costs} as the positional error between the position at any given timestamp $t$ and the desired final position $s_p^f$, the quadratic input and the total time of the trajectory squared.
\autoref{eq:ulimits} enforces that the input force does not exceed the maximum force of the thruster. The timestamp $h_\Delta$ in \autoref{eq:hlimit} must be limited, otherwise the solver fails to find a solution. \autoref{eq:quatnorm} ensures that $s_q$ has unit length and \autoref{eq:quatmult} yields the derivative of the quaternion part of the state by hamiltonian multiplication of the angular part of $v$ and the orientation.\\
\autoref{eq:posint} and \autoref{eq:velint} determine the discrete integration of the state and the velocities. The acceleration is chosen according to the dynamics of the system by the constraint in \autoref{eq:maneqn} according to the equation of motion from the previous chapter. \autoref{eq:psi}, \autoref{eq:psid}, \autoref{eq:phi} and \autoref{eq:phid} define the maximum angles and angular speeds of the thruster. This constraint allows us to ignore the parallel kinematics and work with joints in the simulation even without identifying the actual mass and inertia of the thruster, as the angular velocities and the angle can be used as inputs to the thruster on the physical system. The latter equations define that the first and final states must equal a user-defined target and goal.

\subsection{Trajectory Stabilization}
To compensate for errors in the model, unknown environmental forces and errors in the state estimation, a time-varying LQR stabilizes the \textit{AUV} around the reference trajectory. The state of the system and the control inputs are thus given by:
\begin{eqnarray}
 \hat{x} = x - x^*\label{eq:localx}\\
 \hat{u} = u - u^* \label{eq:localu}
\end{eqnarray}
where $x^*$ and $u^*$ are the optimal state and input given by the trajectory. The linearization is defined as a first-order Taylor Expansion
\begin{equation}
    f(x,u) \approx f(x^*,u^*) + \cfrac{\delta f(s,u)}{\delta s}\hat{x} + \cfrac{\delta f(s,u)}{\delta u}\hat{u}
\end{equation} 
This equation can be written in state space form as
\begin{equation}
    f(x,u) = Ax + Bu
\end{equation}
where the A matrix is given as
\begin{equation}
    A(x) =
    \left( \begin{array}{ccc}
        0^{4\times9} & {\cfrac{\delta q}{\delta w}}^{4\times3} & 0^{4\times5}         \\[8pt]
        0^{5\times9} & 0^{5\times3}                          & \mathbb{I}^{5\times5}\\[3pt]
        \mu^{8\times9}& & C(s,v)^{8\times8}
    \end{array} \right) \label{eq:Amatrix}
\end{equation}
where $\mu$ is:
\begin{equation}
    \mu = M^{-1}\left( \cfrac{\delta r(s)}{\delta s} + \cfrac{\delta B}{\delta s} + \cfrac{\delta C(s,v)}{\delta s} \right)
\end{equation}
and the input matrix as:
\begin{equation*}
B = \\[10pt]
\end{equation*}
\begin{equation}
    \left(\begin{array}{ccc}
      0^{9\times1}&0^{9\times1}&0^{9\times1}  \\
        M^{-1}J^T & M^{-1}\left(\begin{array}{c}
             0^{6\times1} \\ 1 \\ 0 
        \end{array}\right) & M^{-1} \left(\begin{array}{c}
             0^{7\times1} \\ 1
        \end{array}\right)
    \end{array}\right)\label{eq:Bmatrix} 
\end{equation}
Calculating these matrices for all the timestamps where each state is associated with a point in time and then interpolating yields the time-varying matrices $A(t)$ and $B(t)$. Since there is no restriction for LQRs that the state and input matrices must be time-invariant, a time-varying LQR can be computed from $A(t)$ and $B(t)$.\\
Stability describes the dynamics of a system over an infinite time horizon, but our trajectory is only defined over a finite horizon. To stabilize around the trajectory we first have to compute an infinite horizon LQR with the state and input matrix at the last point of the trajectory.
Solving the algebraic Riccati equation:
\begin{equation}
    0 = S^{\infty}A + A^TS^{\infty} - S^{\infty}BR^{-1}B^TS^{\infty}+Q
\end{equation}
yields the cost to go matrix $S$ and the control law:
\begin{equation}
    u^* = -R^{-1}B^TS^{\infty}x = -K^{\infty}x
\end{equation}
The finite horizon LQR can be obtained by solving the differential Riccati Equation
\begin{equation}
    -\dot{S}(t) = S(t)A + A^TS(t)-S(t)BR^{-1}B^TS(t)+Q
\end{equation}
with the terminal condition:
\begin{equation}
    S(T d_{\Delta}) = S^{\infty}
\end{equation}
and the control law:
\begin{equation}
    u^* = -R^{-1}B^TS(t)x = -Kx
\end{equation}
The control law for the complete trajectory can then be written as:
\begin{equation}
    K^f = \left\{
        \begin{array}{ll}
        K  & t \leq T d_{\Delta} \\
        K^{\infty} & \, \textrm{else} \\
        \end{array}\right.
\end{equation}
In this formulation, it can be clearly seen that the linear system is not controllable due to the quaternions as 4 numbers are used to represent 3 rotational DoFs. Similar to \cite{vyaseurognc}, the unit length of the quaternion is therefore enforced directly by reducing the state vector of its first entry. The state size is reduced by one compared to the formulation above. Instead, the real part of the quaternion is calculated as:
\begin{equation}
    q_w = 1 - \sqrt{q_x^2 + q_y^2 + q_z^2}
\end{equation}

\section{Results}
This section will first introduce three trajectories yielded from the Trajectory Optimization. They represent movements close to the boundaries of the mechanical constraints of \textit{DeepLeng} and they heavily rely on hydrodynamics.
The time-varying LQR is tested by executing the reference trajectories in a simulation with the same parameters for hydrodynamic damping used in the Trajectory Optimization.

\subsection{Trajectories}
The following three trajectories are the case study to evaluate the control framework:
(1) Polebalancing, (2) Quarterhelix and (3) Steep elevation. Throughout these cases, we select the maximum and minimum values of the thruster $F_{min}=-\SI{70}{\newton}$ and $F_{max}=\SI{70}{\newton}$.

\begin{figure}[htpb]
\centering
\begin{subfigure}{0.48\linewidth}
  \centering
  \includegraphics[width=\linewidth]{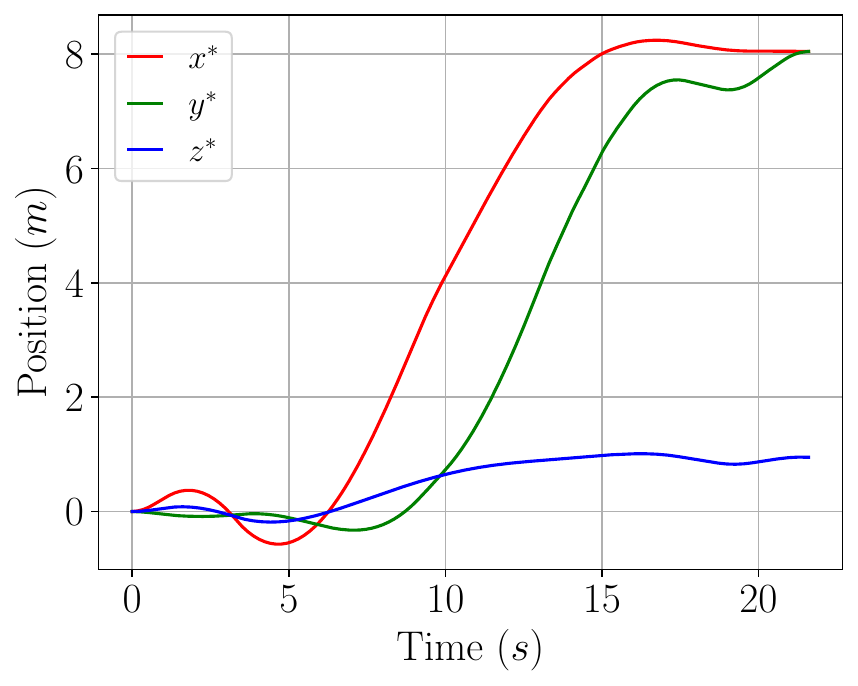}
\end{subfigure}
\hfill
\begin{subfigure}{0.48\linewidth}
  \centering
  \includegraphics[width=\linewidth]{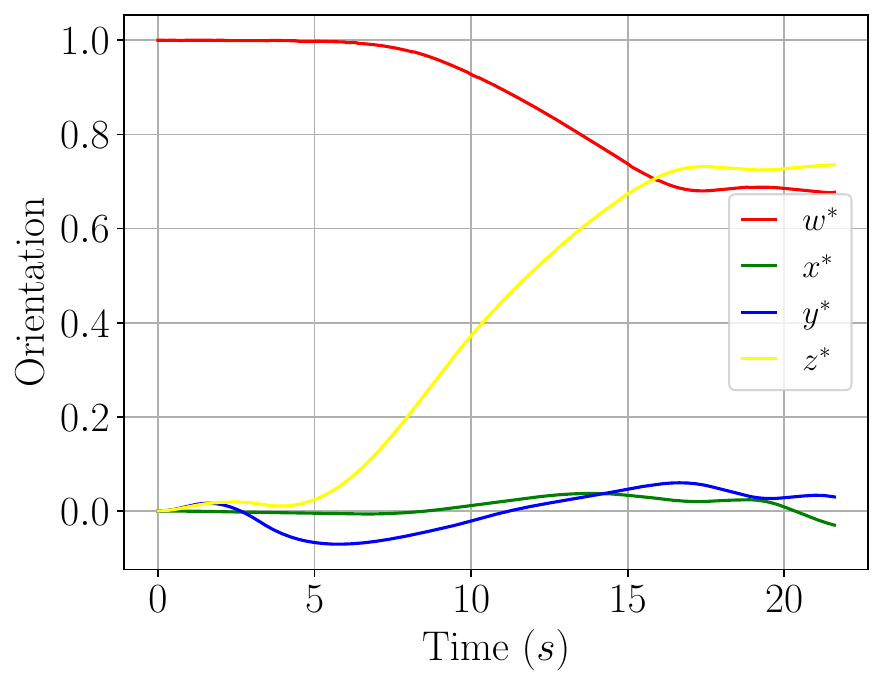}
  \end{subfigure}
\caption{The position and orientation of the quarterhelix trajectory}
\label{fig:quarterhelix_traj}
\end{figure}

\begin{figure}[htpb]
\centering
\begin{subfigure}{0.48\linewidth}
  \centering
  \includegraphics[width=\linewidth]{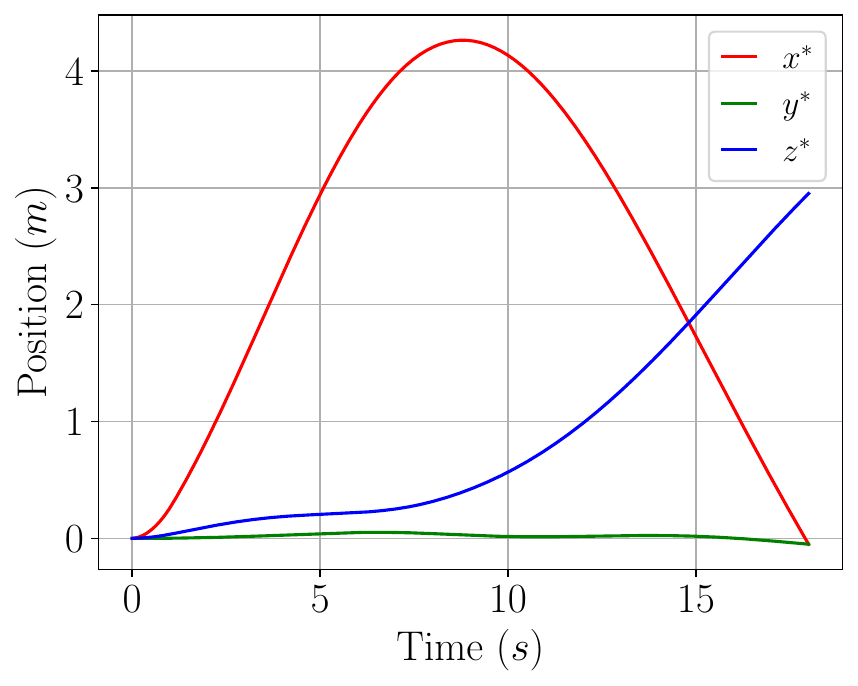}
\end{subfigure}
\hfill
\begin{subfigure}{0.48\linewidth}
  \centering
  \includegraphics[width=\linewidth]{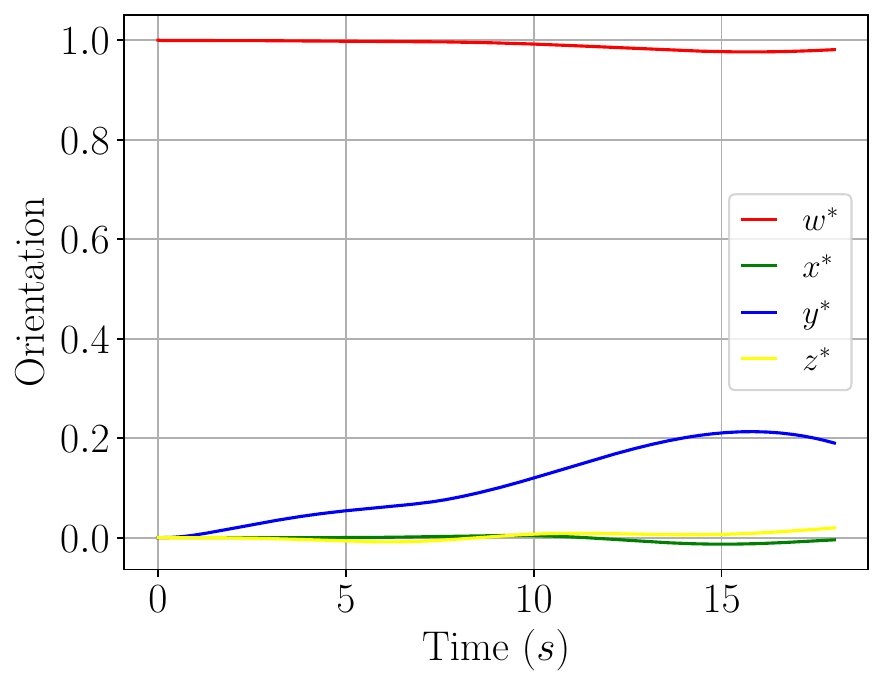}
  \end{subfigure}
\caption{The position and orientation of the steep elevation trajectory}
\label{fig:steepelevation_traj}
\end{figure}

 During the pole balancing maneuver, the vehicle swings up and holds an orientation close to $90^\circ$.  The maneuver is of interest for vertical docking or the inspection of vertical obstacles. It is also similar to the inverted pendulum system.
\begin{figure}[htpb]
\centering
\begin{subfigure}{0.48\linewidth}
  \centering
  \includegraphics[width=\linewidth]{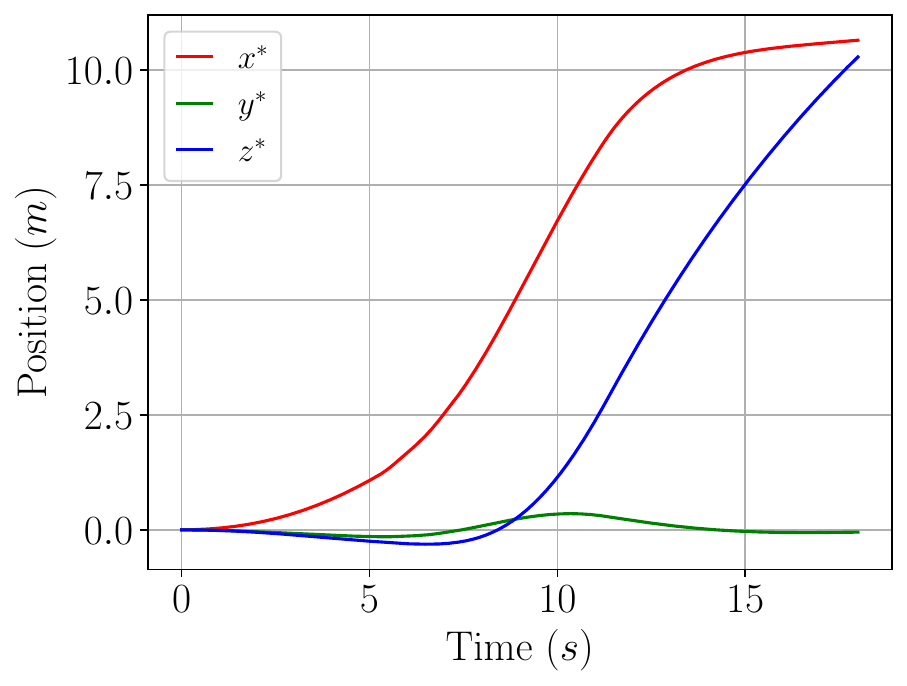}
\end{subfigure}
\hfill
\begin{subfigure}{0.48\linewidth}
  \centering
  \includegraphics[width=\linewidth]{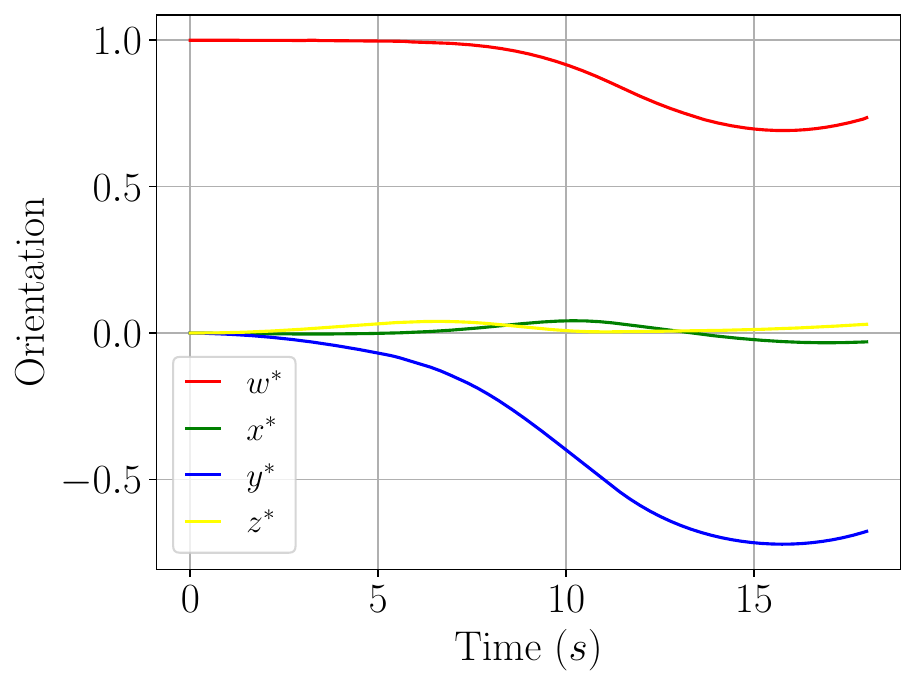}
  \end{subfigure}
\caption{The position and orientation of the pole balancing trajectory}
\label{fig:polebalancing_traj}
\end{figure}
\autoref{fig:polebalancing_traj} shows the reference position and orientation of the polebalancing trajectory. The force limit defined in \autoref{eq:ulimits} was increased for this trajectory, otherwise the restoring torque, which reaches its maximum at a vertical pose, can not be negated by the thrust force. There is still a margin in the maximum angle and angular velocity of the thrusters to allow the time-varying LQR to stabilize around the trajectory.\\
Moving upwards or downwards in a helix is of interest for reaching the seabed of the surface whilst staying in a limited area. The maneuver can be subdivided into quarterhelices, where only a curve with a total rotation of $90^\circ$ is considered. A maneuver of arbitrary duration and range can be easily synthesized by concatenating multiple quarterhelices. \autoref{fig:quarterhelix_traj} shows the trajectories reference position and orientation.\\
For a monotonic elevation in a tight space, hence with constraints in the x and y direction of the \textit{AUV}, the steep elevation trajectory in \autoref{fig:steepelevation_traj} functions as a reference. In emergency situations, it is a fast way to reach the surface.
\subsection{Simulation}
\begin{figure}[ht!]
\centering
\begin{subfigure}{0.48\linewidth}
  \centering
  \includegraphics[width=\linewidth]{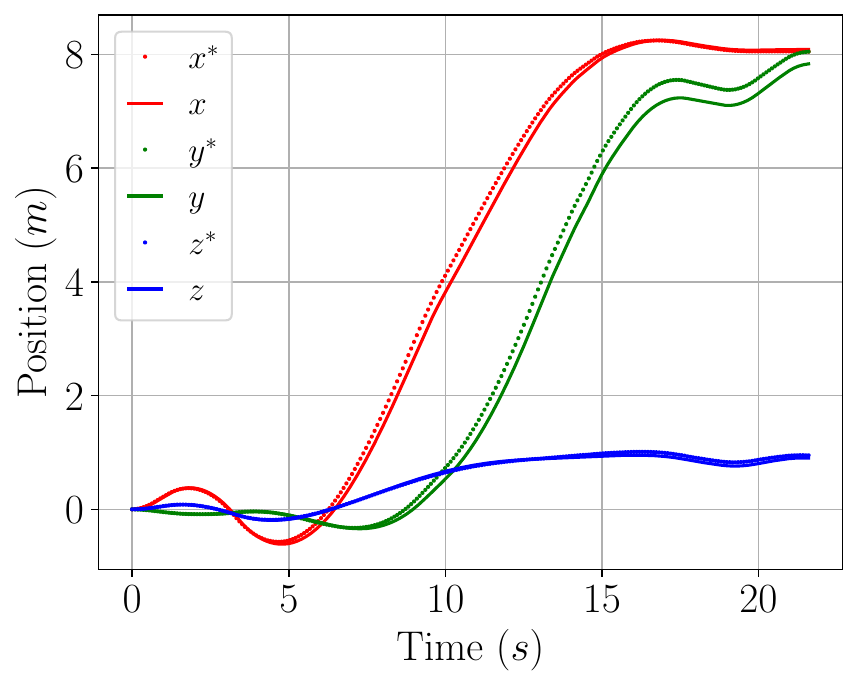}
\end{subfigure}
\hfill
\begin{subfigure}{0.48\linewidth}
  \centering
  \includegraphics[width=\linewidth]{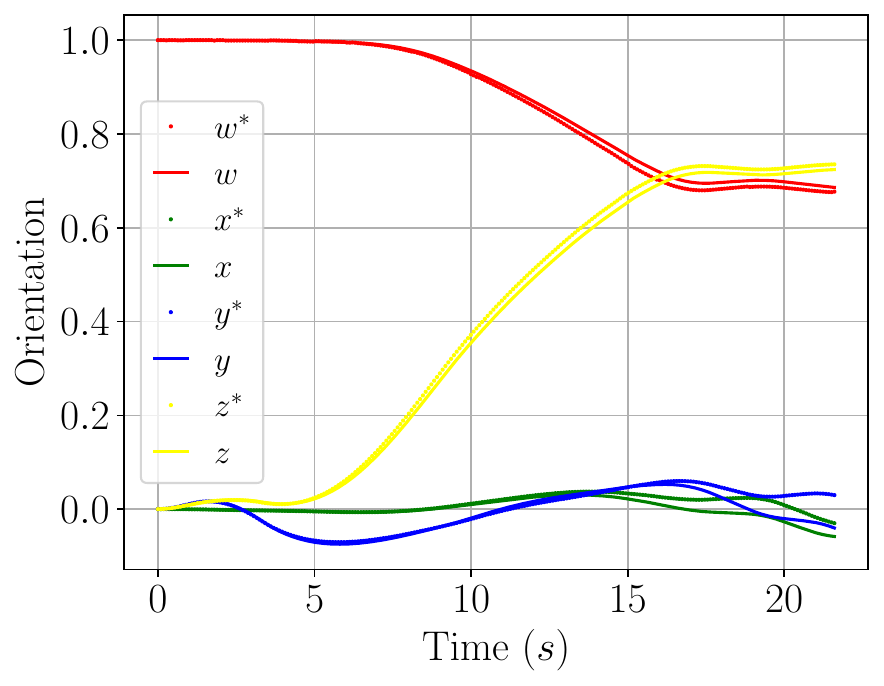}
  \end{subfigure}
\begin{subfigure}{0.48\linewidth}
  \centering
  \includegraphics[width=\linewidth]{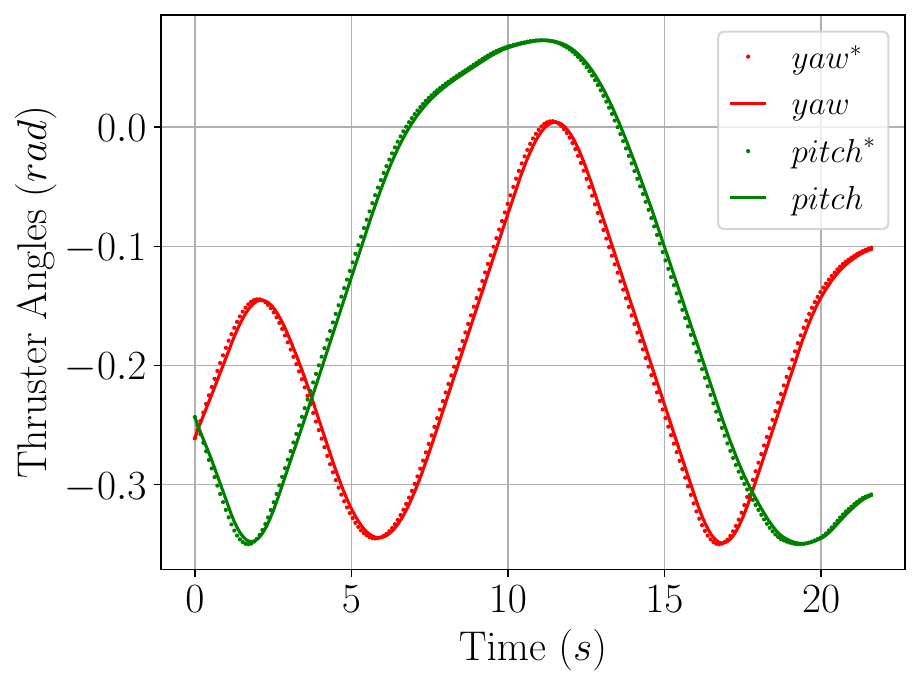}
\end{subfigure}
\hfill
\begin{subfigure}{0.48\linewidth}
  \centering
  \includegraphics[width=\linewidth]{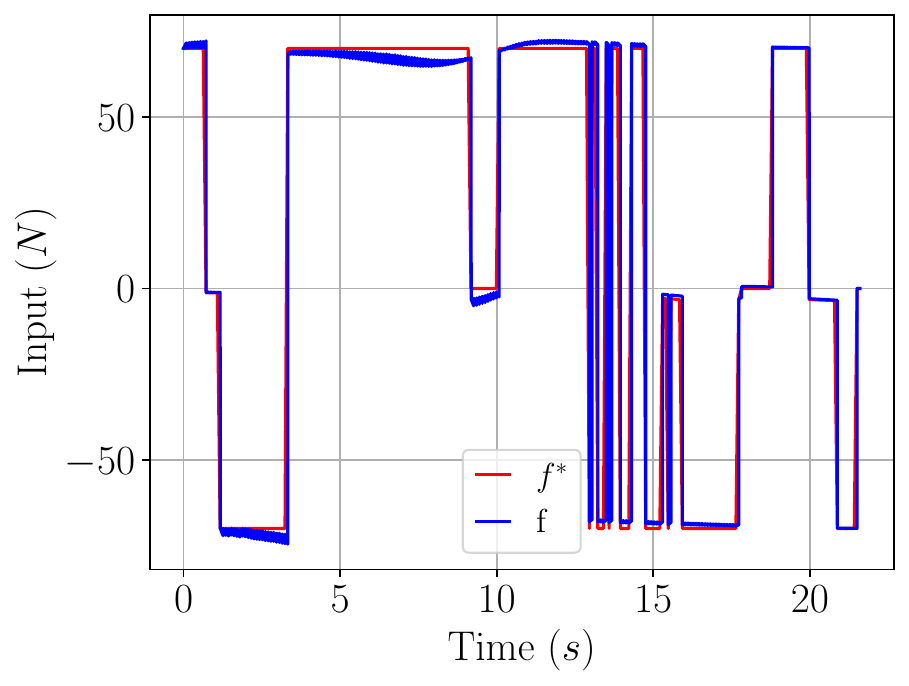}
  \end{subfigure}
\caption{The position(top left), orientation(top right), thruster angles(bottom left) and force input(bottom right) during execution of the quarterhelix trajectory}
\label{fig:quarterhelix}
\end{figure}

\begin{figure}[htpb!]
\centering
\begin{subfigure}{0.48\linewidth}
  \centering
  \includegraphics[width=\linewidth]{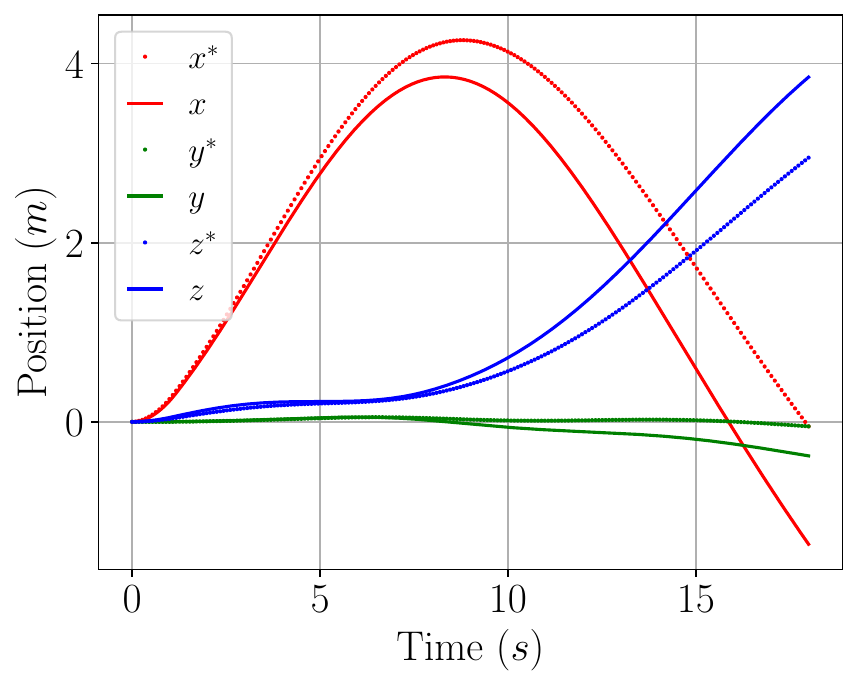}
\end{subfigure}
\hfill
\begin{subfigure}{0.48\linewidth}
  \centering
  \includegraphics[width=\linewidth]{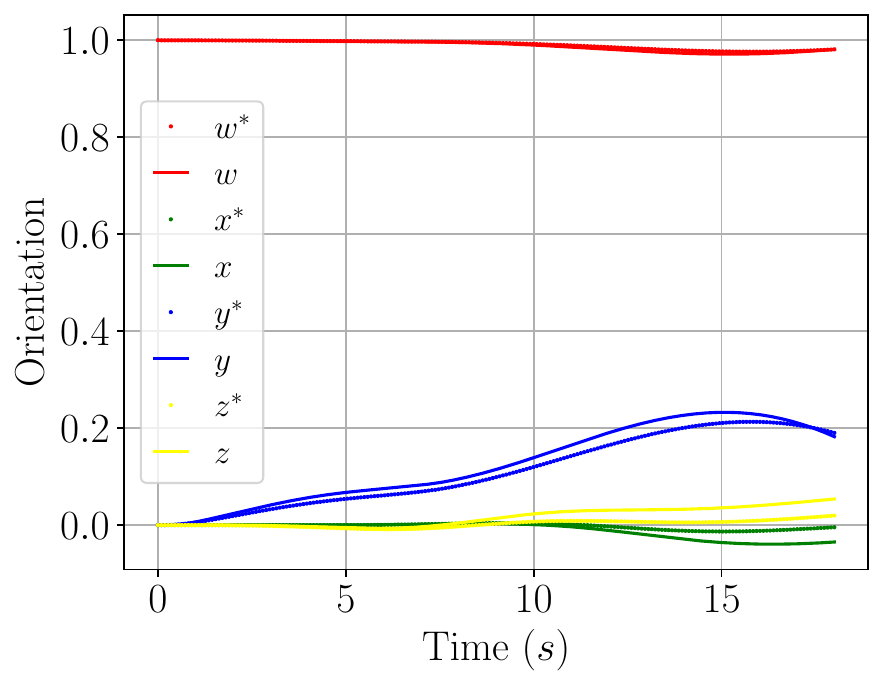}
\end{subfigure}

\begin{subfigure}{0.48\linewidth}
  \centering
  \includegraphics[width=\linewidth]{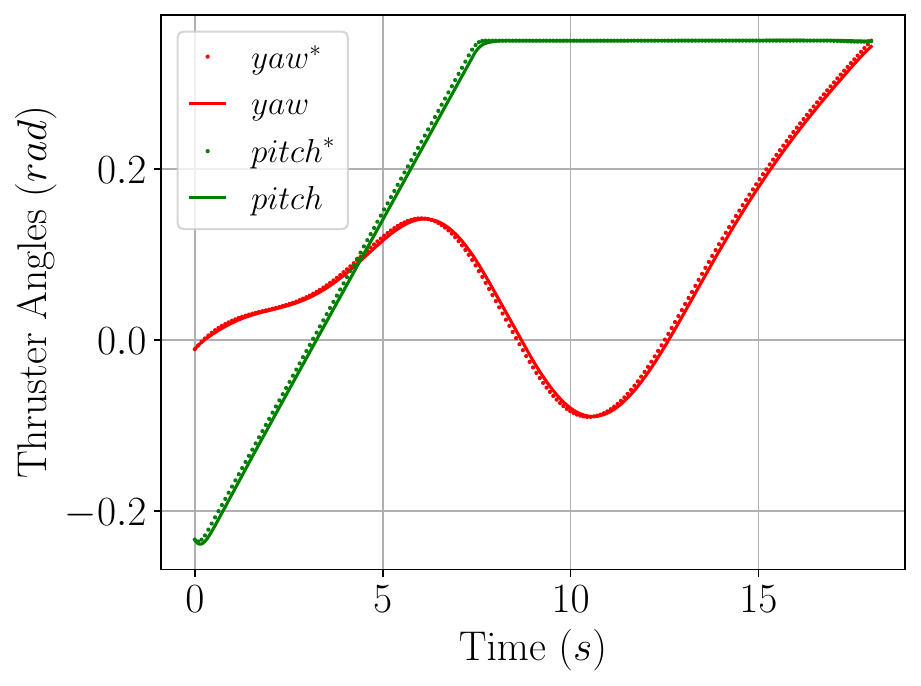}
\end{subfigure}
\hfill
\begin{subfigure}{0.48\linewidth}
  \centering
  \includegraphics[width=\linewidth]{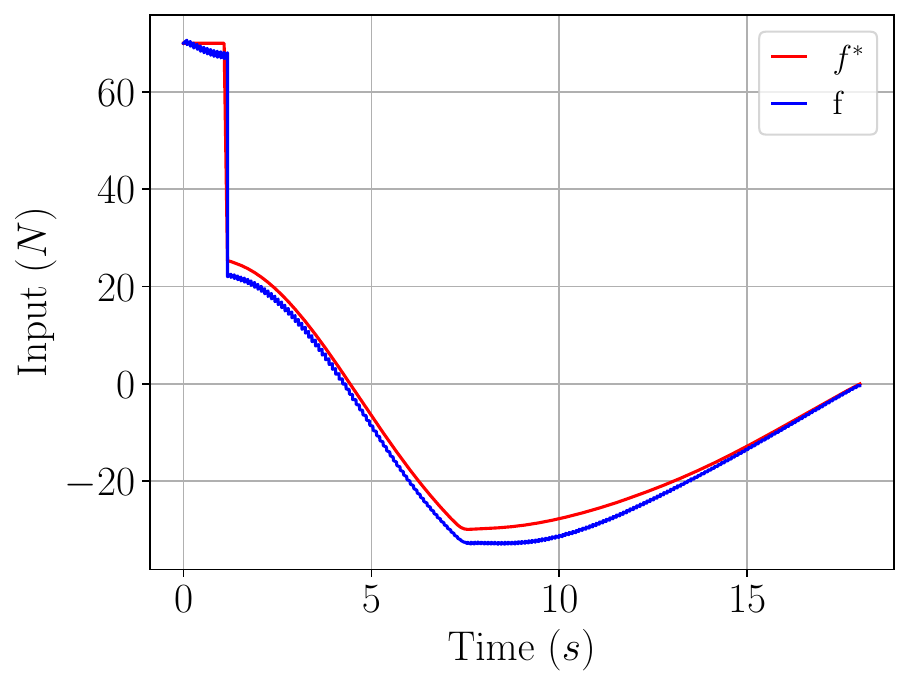}
  \end{subfigure}
\caption{The position(top left), orientation(top right), thruster angles(bottom left) and force input(bottom right) during execution of the steep elevation trajectory}
\label{fig:steepelevation}
\end{figure}

The combined framework of the reference trajectories and the additional time-varying LQR are tested in this simulation. Note that the error passed to the time-varying LQR is not the state of the system, but its difference to the expected state from the reference trajectory. The reference input of the trajectory is subtracted as well.\\
All tests were conducted using the \textit{Drake} simulator. The hydrodynamics were incorporated by implementing a custom module that applies external forces to the \textit{AUV}. It uses the formulation from the System Dynamics chapter.
\begin{figure}[htpb]
\centering
\begin{subfigure}{0.48\linewidth}
  \centering
  \includegraphics[width=\linewidth]{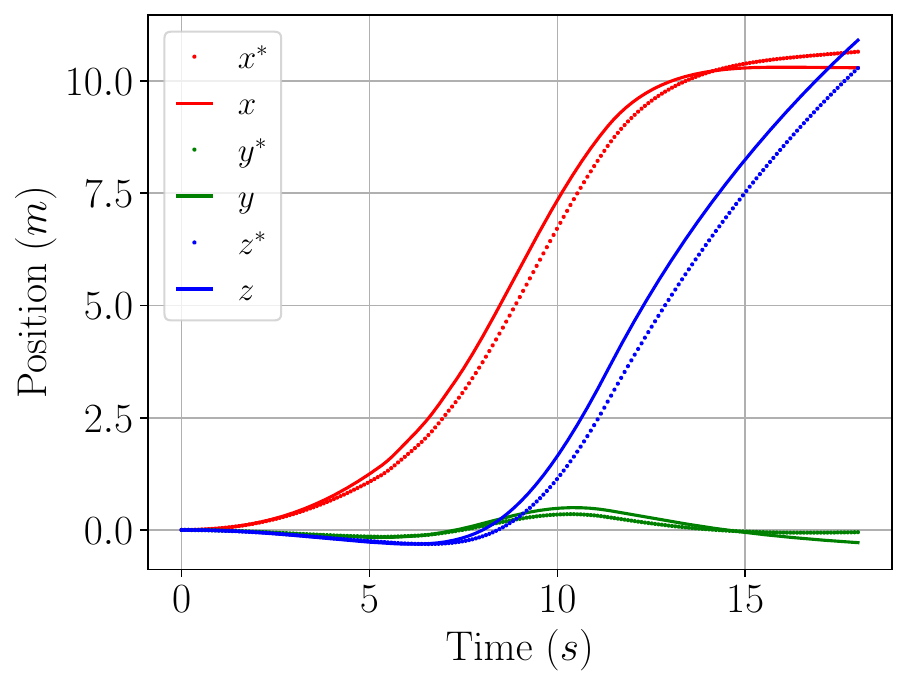}
\end{subfigure}
\hfill
\begin{subfigure}{0.48\linewidth}
  \centering
  \includegraphics[width=\linewidth]{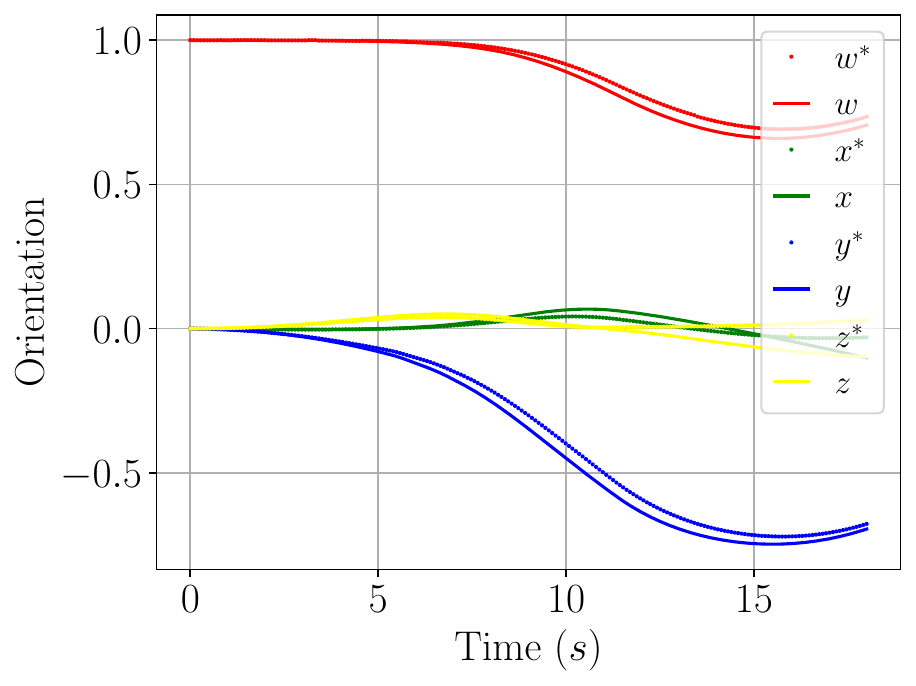}
  \end{subfigure}

\begin{subfigure}{0.48\linewidth}
  \centering
  \includegraphics[width=\linewidth]{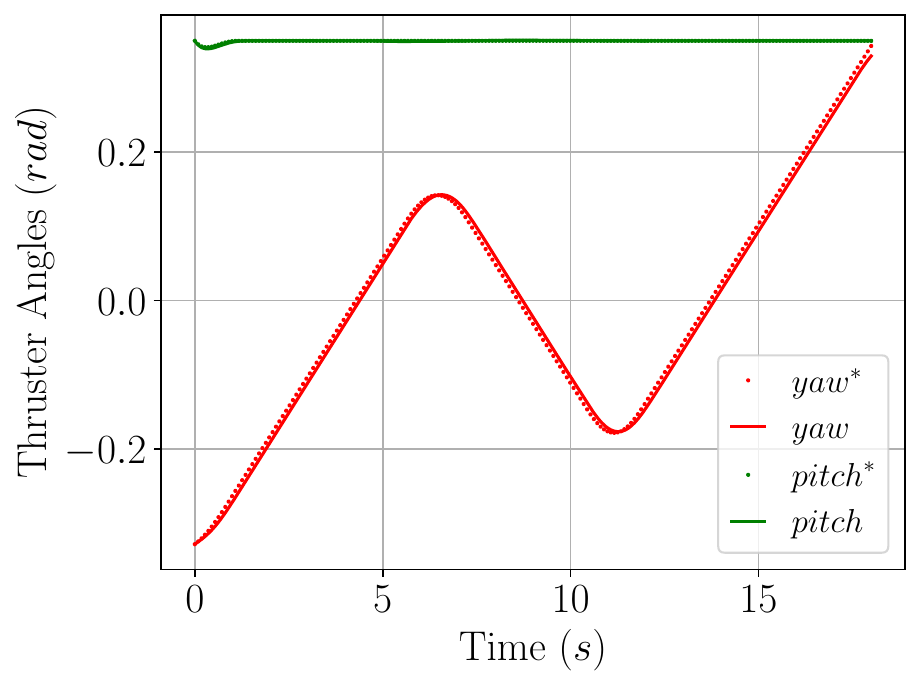}
\end{subfigure}
\hfill
\begin{subfigure}{0.48\linewidth}
  \centering
  \includegraphics[width=\linewidth]{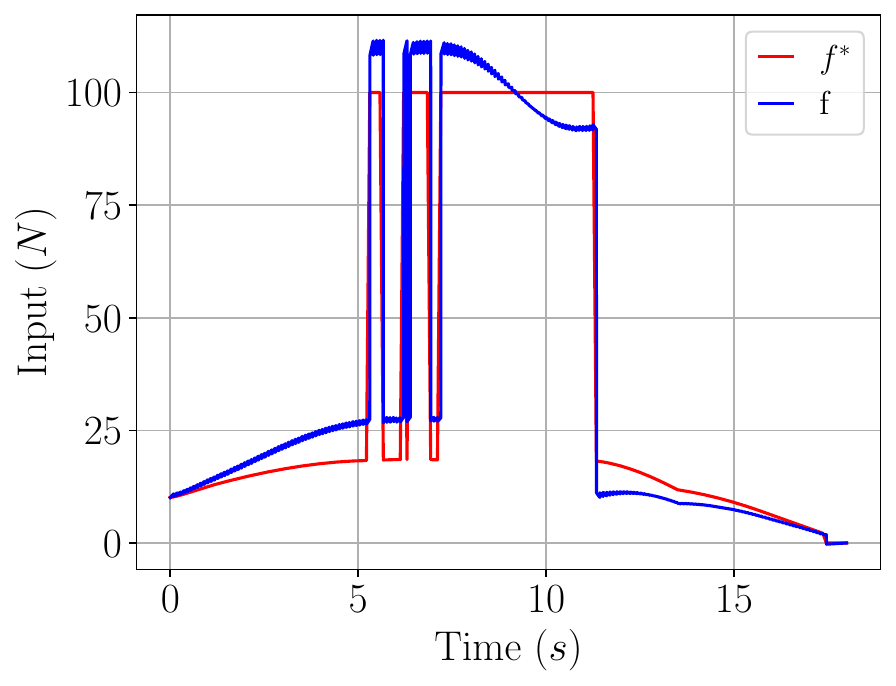}
  \end{subfigure}
\caption{The position(top left), orientation(top right), thruster angles(bottom left) and force input(bottom right) during execution of the pole balancing trajectory}
\label{fig:polebalancing}
\end{figure}
\autoref{fig:polebalancing} shows the position, orientation, the thruster angles and the force input in comparison to the reference trajectories during the execution of the polebalancing trajectory. \autoref{fig:quarterhelix} shows the respective plot for the quarterhelix trajectory and \autoref{fig:steepelevation} for the steep elevation trajectory.

\section{Discussion}
The system matrix $A$ of the time-varying LQR, which models the dynamics of the system, does not contain the hydrodynamic terms that are applied in the Trajectory Optimization and the simulation. Updating it with the additional bias terms would enhance the controller's performance.\\
The representation of the \textit{DeepLeng} is generated by parsing a URDF. The modeling of added mass as proposed by \cite{fossen2011handbook} is not natively supported in URDF, since that would require the possibility to specify masses for each axis independently. For the above simulation, an average of the added mass was added to the vehicle's mass in the URDF to approximate the dynamics.\\
\cite{tedrake2016underactuated} describes how to synthesize a controller that is proven to stabilize around the trajectory. Since stability is considered as a statement when time goes to infinity, not only a finite horizon LQR over the time interval of the trajectory is needed but also an infinite horizon LQR which stabilizes around the final pose of the trajectory. The optimal cost-to-go function $S$ of the latter infinite horizon LQR is passed into the finite horizon LQR as a cost function for the final state. In the simulation of this paper, this leads to the failure of the calculation of the finite horizon LQR. Better tuning of the $Q$ and $R$ matrices of the infinite horizon LQR may lead to a cost-to-go matrix $S$ with smaller values which allows the successful computation of the finite horizon LQR.\\
Till the submission of the paper tuning the time-varying LQR lead to better performance. To the author's assessment, the tuning is still not optimal, and further tuning could improve the performance.
\section{Conclusion}
This paper presented a Trajectory Optimization formulation for an underactuated \textit{AUV} and a time-varying LQR to stabilize around the reference trajectories. Simulations showed the capability of the controller to execute representative trajectories.\\
For icy moon exploration, the proposed control architecture will be able to generate and stabilize around nominal trajectories given a precise model, especially of the hydrodynamics. Changes in temperature and alterations in the composition of the liquid have a severe influence on the hydrodynamic parameters. Since the exact properties of the ocean on \textit{Europa} are unknown and due to the controller's dependency on a good model, model identification on Earth will not provide sufficient results.\\
Adding an online model adaptation could allow the system to correct the errors in the model identification and result in robustness toward unknown dynamics for ocean exploration.

\section*{Acknowledgements}
This work was supported by the project TRIPLE-GNC which received funding from the German Ministry for Economic Affairs and Climate Action (BMWK), FKZ: 50NA2306C.
The second author would like to acknowledge the support of M-RoCk (Grant No.: FKZ 01IW21002) and AAPLE (Grant Number: 50WK2275) and the European Union’s Horizon 2020 research and innovation program under the Marie Skłodowska-Curie grant agreement No 813644.

\bibliography{references}

\end{document}